\documentclass[runningheads,a4paper]{llncs}

\usepackage{llncsdoc}
\usepackage{mathptmx}       
\usepackage{helvet}         
\usepackage{courier}        
\usepackage{type1cm}        
\usepackage{makeidx}         
\usepackage[pdftex]{graphicx}
\DeclareGraphicsExtensions{.pdf,.jpeg,.png}
\usepackage{multicol}        
\usepackage[bottom]{footmisc}
\usepackage{amsmath}
\usepackage{amssymb}
\usepackage{url}
\usepackage[linesnumbered,ruled,vlined]{algorithm2e}
\usepackage{array}
\usepackage{booktabs}
\usepackage[caption=false,font=footnotesize]{subfig}
\usepackage{xcolor,colortbl}

\definecolor{Gray}{gray}{0.85}
\definecolor{LightCyan}{rgb}{0.88,1,1}

\makeindex             

\begin{document}

\mainmatter

\title{Effective Blind Source Separation Based on the Adam Algorithm}
\author{Michele Scarpiniti \and Simone Scardapane \and Danilo Comminiello \and Raffaele Parisi \and Aurelio Uncini}
\titlerunning{Effective Blind Source Separation Based on the Adam Algorithm}
\authorrunning{Scarpiniti et al.}
\institute{Department of Information Engineering, Electronics and Telecommunications (DIET), \\ ``Sapienza'' University of Rome, \\ Via Eudossiana 18, 00184, Rome. \\Email: \{michele.scarpiniti, simone.scardapane, danilo.comminiello, raffaele.parisi, aurelio.uncini\}@uniroma1.it}
%
%
\maketitle

\begin{abstract}
In this paper, we derive a modified InfoMax algorithm for the solution of Blind Signal Separation (BSS) problems by using advanced stochastic methods. The proposed approach is based on a novel stochastic optimization approach known as the Adaptive Moment Estimation (Adam) algorithm. The proposed BSS solution can benefit from the excellent properties of the Adam approach. In order to derive the new learning rule, the Adam algorithm is introduced in the derivation of the cost function maximization in the standard InfoMax algorithm. The natural gradient adaptation is also considered. Finally, some experimental results show the effectiveness of the proposed approach.
\end{abstract}

\section{Introduction}
\label{sec:intro}

Blind Source Separation (BSS) is a well-known and well-studied field in the adaptive signal processing and machine learning \cite{cichocki2002,Comon2010,Haykin2000,Choi2005,Scarpiniti2008,Vigliano2008}. The problem is to recover original and unknown sources from a set of mixtures recorded in an unknown environment. The term \textit{blind} refers to the fact that both the sources and the mixing environment are unknown. 

Several well-performing approaches exist when the mixing environment is instantaneous \cite{Comon2010,Bell1995a}, while some problems still arise in convolutive environments \cite{Araki2003,Boulmezaoud2010,Smaragdis1998}. Different approaches were proposed to solve BSS in linear and instantaneous environment. Some of these approaches perform separation by using high order statistics (HOS) while others exploit information theoretic (IT) measures \cite{cichocki2002}. One of the well-know algorithms in this latter class is the InfoMax one proposed by Bell and Sejnowski in \cite{Bell1995a}. The InfoMax algorithm is based on the maximization of the joint entropy of the output of a single layer neural network and it is very efficient and easy to implement since the gradient of the joint entropy can be evaluated simply in a closed form. Moreover, in order to avoid numerical instability, a natural gradient modification to InfoMax algorithm has also been proposed \cite{Amari1998,cichocki2002}.

Unfortunately, all these solutions perform slowly when the number of the original sources to be separated is high and/or bad scaled. The separation becomes impossible if the number of sources is equal or greater than ten. In addition, the convergence speed problem worsen in the case of additive sensor noise to mixtures or when the mixing matrix is close to be ill-conditioned. However, specially when working with speech and audio signals, fast convergence speed is an important task to be performed. Many authors have tried to overcome this problem: some solutions consist in incorporating a momentum term in the learning rule \cite{Liu2009}, in a self-adjusting variable step-size \cite{Thomas2000} or in a scaled natural gradient algorithm \cite{Douglas2007}.

Recently, a novel algorithm for gradient based optimization of stochastic cost functions has been proposed by Kingma and Ba in \cite{Kingma2015}. This algorithm is based on the adaptive estimates of the first and second order moments of the gradient, and for this reason has been called the Adaptive Moment Estimation (Adam) algorithm. The authors have demonstrated in \cite{Kingma2015} that Adam is easy to implement, computationally efficient, invariant to diagonal rescaling of the gradients and well suited for problems with large data and parameters.

The Adam algorithm combines the advantages of other state-of-the-art optimization algorithms, like AdaGrad \cite{Duchi2011} and RMSProp \cite{Tieleman2012}, outperforming the limitations of these algorithms. In addition, Adam can be related to the natural gradient (NG) adaptation \cite{Amari1998}, employing a preconditioning that adapts to the geometry of data.

In this paper we propose a modified InfoMax algorithm based on the Adam algorithm \cite{Kingma2015} for the solution of BSS problems. We derive the proposed modified algorithm by using the Adam algorithm instead of the standard stochastic gradient ascent rule. It is shown that the novel algorithm has a faster convergence speed with respect to the standard InfoMax algorithm and usually also reaches a better separation. Some experimental results, evaluated in terms of the Amari Performance index (PI) \cite{cichocki2002}, show the effectiveness of the proposed idea.

The rest of the paper is organized as follows. In Section \ref{sec:bss} we briefly introduce the problem of BSS. Then, we give some details on the Adam algorithm in Section \ref{sec:adam}. The main novelty of this paper, the extension of InfoMax algorithm with Adam is provided in Section \ref{sec:adam_InfoMax}. Finally, we validate our approach in Section \ref{sec:results}. We conclude with some final remarks in Section \ref{sec:conclusions}.

\section{The Blind Source Separation Problem}
\label{sec:bss}
Let us consider a set of $N$ unknown and statistically independent sources denoted as ${\bf{s}}[n] = \left[ s_1[n], \ldots, s_N[n] \right]^T$, such that the components $s_i[n]$ are zero-mean and mutually independent. Signals received by an array of $M$ sensors are denoted by ${\bf{x}}[n] = \left[ x_1[n], \ldots, x_M[n] \right]^T$ and are called mixtures. For simplicity, here we consider the case of $N=M$.

In the case of a linear and instantaneous mixing environment, the mixture can be described in a matrix form as
\begin{equation}
{\bf{x}}[n] = {\bf{A}} {\bf{s}}[n] + {\bf{v}}[n],
\label{eq:mixtures}
\end{equation}
where the matrix ${\bf{A}} = \left( a_{ij} \right)$ collects the mixing coefficients $a_{ij}$, $i,j = 1,2,\ldots,N$ and ${\bf{v}}[n] = \left[ v_1[n], \ldots, v_N[n] \right]^T$ is a noise vector, with correlation matrix ${\bf{R}}_{\bf{v}} = \sigma^2_v {\bf{I}}$ and noise variance $\sigma^2_v$.

The separated signals ${\bf{u}}[n]$ are obtained by a separating matrix ${\bf{W}} = \left( w_{ij} \right)$ as described by the following equation
\begin{equation}
{\bf{u}}[n] = {\bf{W}} {\bf{x}}[n].
\label{eq:estimates}
\end{equation}
The transformation in \eqref{eq:estimates} is such that ${\bf{u}}[n] = \left[ u_1[n], \ldots, u_N[n] \right]^T$ has components $u_k[n]$, $k = 1,2,\ldots,N$, that are as independent as possible.

Moreover, due to the well-known permutation and scaling ambiguity of the BSS problem, the output signals ${\bf{u}}[n]$ can be expressed as
\begin{equation}
{\bf{u}}[n] = {\bf{PD}} {\bf{s}}[n],
\label{eq:permutation}
\end{equation}
where $\bf{P}$ is an $N \times N$ permutation matrix and $\bf{D}$ is an $N \times N$ diagonal scaling matrix.
\begin{figure}[htb]
\centering
\includegraphics[width=0.8\columnwidth]{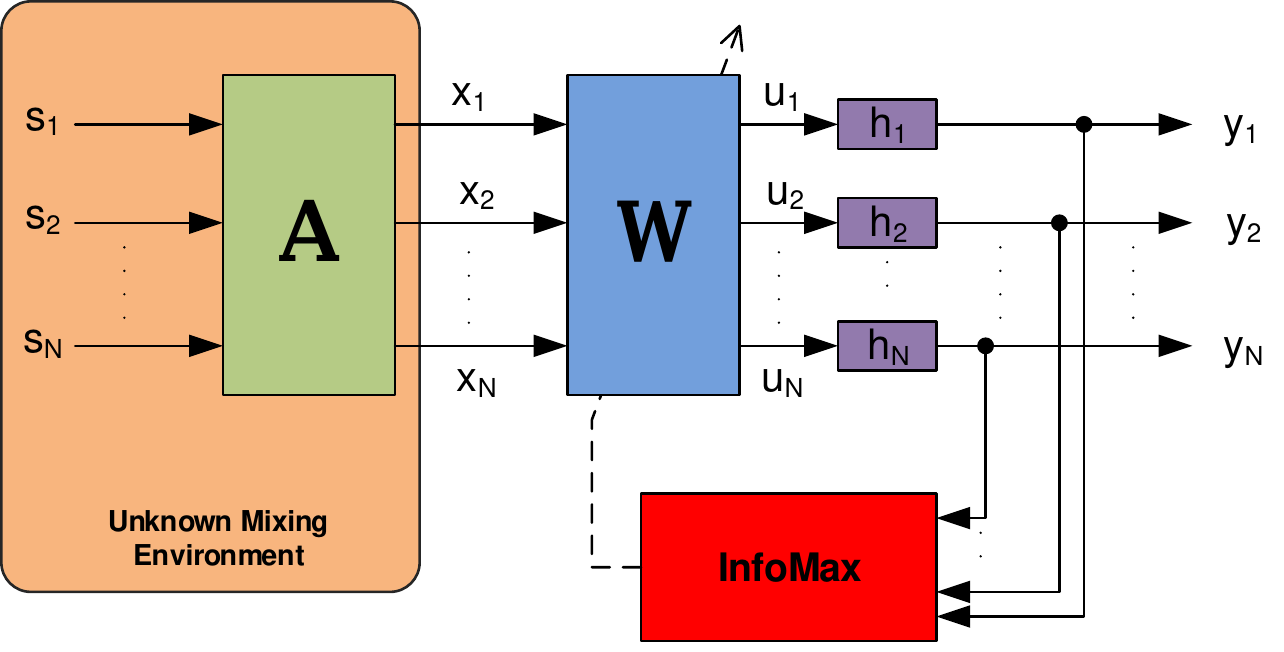}
\caption{Unknown mixing environment and the InfoMax network.}
\label{fig:Infomax}
\end{figure}

\noindent The weights $w_{ij}$ can be adapted by maximizing or minimizing some suitable cost function \cite{cichocki2002,Choi2005}. A particularly good approach is to maximize the joint entropy of a single layer neural network \cite{Bell1995a}, as shown in Figure \ref{fig:Infomax}, leading to the Bell and Sejnowski InfoMax algorithm. In this network each output $y_i[n]$ is a nonlinear transformation of each signal $u_i[n]$: 
\begin{equation}
y_i[n] = h_i\left( u_i[n] \right).
\label{eq:AF}
\end{equation}
Each function $h_i(\cdot)$ is known as activation function (AF).

With reference to Figure \ref{fig:Infomax}, using the equation relating the probability density function (pdf) of a random variable $p_{\bf{x}}\left( {\bf{x}} \right)$ and nonlinear transformation of it $p_{\bf{y}}\left( {\bf{y}} \right)$ \cite{Papoulis1991}, the joint entropy $H\left( {\bf{y}} \right)$ of the network output ${\bf{y}}[n]$ can be evaluated as
\begin{equation}
H\left( {\bf{y}} \right) \equiv - E\left\{ \ln p_{\bf{y}}\left( {\bf{y}} \right) \right\} = H\left( {\bf{x}} \right) + \ln \det {\bf{W}} + \sum_{i=1}^N \ln \left| h'_i \right|,
\label{eq:entropy_1}
\end{equation}
where $h_i'$ is the first derivative of the $i$-th AF with respect its input $u_i[n]$.

Evaluating the gradient of \eqref{eq:entropy_1} with respect to the separating parameters $\bf{W}$, after some not too complicated manipulations, leads to
\begin{equation}
\nabla_{\bf{W}} H\left( {\bf{y}} \right) = {\bf{W}}^{-T} + {\boldsymbol{\Psi}} {\bf{x}}^T[n],
\label{eq:entropy_gradient}
\end{equation}
where $(\cdot)^{-T}$ denotes the transpose of the inverse and ${\boldsymbol{\Psi}} = \left[ \Psi_1, \ldots, \Psi_N \right]^T$ is a vector collecting the terms $\Psi_k = h_k''/h_k'$, defined as the ratio of the second and the first derivatives of the AFs.

In order to avoid the numerical problems of the matrix inversion in \eqref{eq:entropy_gradient} and the possibility to remain blocked in a local minimum, Amari has introduced the Natural Gradient (NG) adaptation \cite{Amari1998} that overcomes such problems. The NG adaptation rule, can be obtained simply by right multiplying the stochastic gradient for the term ${\bf{W}}^T{\bf{W}}$. Hence, after multiplying \eqref{eq:entropy_gradient} for this term, the NG InfoMax is simply
\begin{equation}
\nabla_{\bf{W}} H\left( {\bf{y}} \right) = \left( {\bf{I}} + {\boldsymbol{\Psi}} {\bf{u}}^T[n] \right) {\bf{W}}.
\label{eq:entropy_natural_gradient}
\end{equation}
Regarding the selection of the AFs shape, there are several alternatives. However, especially in the case of audio and speech signals, a good nonlinearity is represented by the $\tanh(\cdot)$ function. With this choice, the vector $\boldsymbol{\Psi}$ in \eqref{eq:entropy_gradient} and \eqref{eq:entropy_natural_gradient} is simply evaluated as $-2 {\bold{y}}[n]$.

In summary, using the $\tanh(\cdot)$ AF, the InfoMax and natural gradient InfoMax algorithms are described by the following learning rules:
\begin{align}
{\bf{W}}_t &= {\bf{W}}_{t-1} + \mu \left( {\bf{W}}_{t-1}^{-T} - {\bf{y}}[n] {\bf{x}}^T[n] \right), \label{eq:InfoMax} \\
{\bf{W}}_t &= {\bf{W}}_{t-1} + \mu \left( {\bf{I}} - {\bf{y}}[n] {\bf{u}}^T[n] \right) {\bf{W}}_{t-1}, \label{NG_InfoMax}
\end{align}
where $\mu$ is the learning rate or step-size.

\section{The Adam Algorithm}
\label{sec:adam}
Le us denote with $f\left( \boldsymbol{\theta} \right)$ a noisy cost function to be minimized (or maximized) with respect to the parameters $\boldsymbol{\theta}$. The problem is considered stochastic for the random nature of data samples or for inherent function noise. In the following, the noisy gradient vector at time $t$ of the cost function $f\left( \boldsymbol{\theta} \right)$ with respect to the parameters $\boldsymbol{\theta}$, will be denoted with ${\bf{g}}_t = \nabla_{\boldsymbol{\theta}} f_t \left( \boldsymbol{\theta} \right)$.

The Adam algorithm performs the gradient descent (or ascent) optimization by evaluating the moving averages of the noisy gradient ${\bf{m}}_t$ and the square gradient ${\bf{v}}_t$ \cite{Kingma2015}. These moment vectors are updated by using two scalar coefficients $\beta_1$ and $\beta_2$ that control the exponential decay rates:
\begin{align}
{\bf{m}}_t &= \beta_1 {\bf{m}}_{t-1} + \left( 1 - \beta_1 \right) {\bf{g}}_t, \label{eq:m_update} \\
{\bf{v}}_t &= \beta_2 {\bf{v}}_{t-1} + \left( 1 - \beta_2 \right) {\bf{g}}_t \odot {\bf{g}}_t, \label{eq:v_update} 
\end{align}
where $\beta_1, \beta_2 \in \left[0, 1\right)$ and $\odot$ denotes the element-wise multiplication, while ${\bf{m}}_0$ and ${\bf{v}}_0$ are initialized as zero vectors. These vectors represent the mean and the uncentered variance of the gradient vector ${\bf{g}}_t$. Since the estimates of ${\bf{m}}_t$ and ${\bf{v}}_t$ are biased towards zero, due to their initialization, a bias correction is computed on these moments
\begin{align}
\hat{\bf{m}}_t &= \frac{{\bf{m}}_t}{1 - \beta_1^t}, \label{eq:m_bias} \\
\hat{\bf{v}}_t &= \frac{{\bf{v}}_t}{1 - \beta_2^t}. \label{eq:v_bias} 
\end{align}
The vector $\hat{\bf{v}}_t$ represents an approximation of the diagonal of the Fisher information matrix \cite{Pascanu2014}. Hence Adam can be related to the natural gradient algorithm \cite{Amari1998}.

Finally, the parameter vector ${\boldsymbol{\theta}}_t$ at time $t$, is updated by the following rule
\begin{equation}
{\boldsymbol{\theta}}_t = {\boldsymbol{\theta}}_{t-1} - \eta \frac{\hat{\bf{m}}_t}{\sqrt{\hat{\bf{v}}_t} + \varepsilon},
\label{eq:param_update}
\end{equation}
where $\eta$ is the step size and $\varepsilon$ is a small positive constant used to avoid the division for zero. In the gradient ascent, the minus sign in \eqref{eq:param_update} is substituted with the plus sign.

\section{Modified InfoMax algorithm}
\label{sec:adam_InfoMax}
In this section, we introduce the modified Bell and Sejnowski InfoMax algorithm, based on the Adam optimization method. Since Adam algorithm uses a vector of parameters, we perform a vectorization of the gradient \eqref{eq:entropy_gradient} or \eqref{eq:entropy_natural_gradient}
\begin{equation}
{\bf{w}} = {\mathrm{vec}}\left( \nabla_{\bf{W}} H\left( {\bf{y}} \right) \right) \in \mathbb{R}^{N^2 \times 1},
\label{eq:grad_vec}
\end{equation}
where ${\mathrm{vec}}\left( {\bf{A}} \right)$ is the vectorization operator, that forms a vector by stacking the columns of the matrix ${\bf{A}}$ below one another. The gradient vector is evaluated on a number of blocks $N_B$ extracted from the signals.

At this point, the mean and variance vectors are evaluated from the knowledge of the gradient ${\bf{w}}_t$ at time $t$ by using equations \eqref{eq:m_update}--\eqref{eq:v_bias}. Then, using \eqref{eq:param_update}, the gradient vector \eqref{eq:grad_vec} is updated for the maximization of the joint entropy by
\begin{equation}
{\bf{w}}_t = {\bf{w}}_{t-1} + \eta \frac{\hat{\bf{m}}_t}{\sqrt{\hat{\bf{v}}_t} + \varepsilon}.
\label{eq:Adam_Infomax}
\end{equation}
Finally, the vector ${\bf{w}}_t$ is reshaped in matrix form, by
\begin{equation}
{\bf{W}}_t = {\mathrm{mat}}\left( {\bf{w}}_t \right) \in \mathbb{R}^{N \times N},
\label{eq:vec_to_mat}
\end{equation}
where ${\mathrm{mat}}\left( {\bf{a}} \right)$ reconstructs the $N \times N$ matrix by unstacking the columns from the vector $\bf{a}$. The whole algorithm is in case  repeated for a certain number of epochs $N_{ep}$.

The pseudo-code of the modified InfoMax algorithm with Adam, called here Adam InfoMax, is described in Algorithm \ref{alg:Adam_Infomax}.
\begin{algorithm}[htb]
\DontPrintSemicolon
\SetAlgoLined
\KwData{Mixture signals ${\bf{x}}[n]$, $\eta$, $\beta_1$, $\beta_2$, $\varepsilon$, $N_B$, $N_{ep}$.}
\textbf{Initialization:} ${\bf{W}}_0 = {\bf{I}}$, ${\bf{m}}_0 = {\bf{0}}$, ${\bf{v}}_0 = {\bf{0}}$, ${\bf{w}}_0 = {\bf{0}}$\;
  $P = N_B N_{ep}$\;
  \For{t=1:P}{
	Extract the $t$-th block ${\bf{x}}_t$ from ${\bf{x}}[n]$\;
	${\bf{u}}_t = {\bf{W}}_{t-1} {\bf{x}}_t$\;
	${\bf{y}}_t = \tanh\left( {\bf{u}}_t \right)$\;
	Update gradient $\nabla_{\bf{W}} H\left( {\bf{y}}_t \right)$ in \eqref{eq:entropy_gradient} or \eqref{eq:entropy_natural_gradient}\;
	${\bf{w}}_t = {\mathrm{vec}}\left( \nabla_{\bf{W}} H\left( {\bf{y}}_t \right) \right) $\;
	${\bf{m}}_t = \beta_1 {\bf{m}}_{t-1} + \left( 1 - \beta_1 \right) {\bf{w}}_t$\;
  ${\bf{v}}_t = \beta_2 {\bf{v}}_{t-1} + \left( 1 - \beta_2 \right) {\bf{w}}_t \odot {\bf{w}}_t$\;
  $\hat{\bf{m}}_t = \frac{{\bf{m}}_t}{1 - \beta_1^t}$\;
  $\hat{\bf{v}}_t = \frac{{\bf{v}}_t}{1 - \beta_2^t}$\;
	${\bf{w}}_t = {\bf{w}}_{t-1} + \eta \frac{\hat{\bf{m}}_t}{\sqrt{\hat{\bf{v}}_t} + \varepsilon}$\;
	${\bf{W}}_t = {\mathrm{mat}}\left( {\bf{w}}_t \right)$\;
  }
	\KwResult{Separated signals: ${\bf{u}}[n] = {\bf{W}}_P {\bf{x}}[n]$}
\caption{Pseudo-code for the Adam InfoMax algorithm.}
\label{alg:Adam_Infomax}
\end{algorithm}

\section{Experimental Results}
\label{sec:results}
In this section, we propose some experimental results to demonstrate the effectiveness of the proposed idea. We perform separation of mixtures of both synthetic and real-world data. The results are evaluated in terms of the Amari Performance Index (PI) \cite{cichocki2002}, defined as
\begin{equation}
PI = \frac{1}{N\left( N - 1 \right)} \sum_{i = 1}^N \left[ \left( \sum_{k = 1}^N \frac{\left| {q_{ik} } \right|}{\mathop {\max}\limits_j {\left| {q_{ij} } \right|} } - 1 \right) + \left( \sum_{k = 1}^N \frac{\left| {q_{ki} } \right|}{\mathop {\max}\limits_j {\left| {q_{ji} } \right|} } - 1 \right) \right],
\label{eq:PI}
\end{equation}
where $q_{ij}$ are the elements of the matrix ${\bf{Q}} = {\bf{WA}}$. This index is close to zero if the matrix $\bf{Q}$ is close to the product of a permutation matrix and a diagonal scaling matrix.

The performances of the proposed algorithm were also compared with the standard InfoMax algorithm \cite{Bell1995a} and the Momentum InfoMax described in \cite{Liu2009}. In this last algorithm, the $\alpha$ parameter is set to 0.5 in all experiments.

In a first experiment, we perform the separation of five mixtures obtained as linear combination of the following bad-scaled independent sources
\[
\begin{split}
s_1[n] &= 10^{-6} \cdot \sin \left( 350n \right) \sin \left( 60n \right), \\
s_2[n] &= 10^{-5} \cdot {\rm{tri}} \left( 70n \right), \\
s_3[n] &= 10^{-4} \cdot \sin \left( 800n \right) \sin \left( 80n \right), \\
s_4[n] &= 10^{-5} \cdot \cos \left( 400n + 4 \cos \left( 60n \right) \right), \\
s_5[n] &= \xi[n],
\end{split}
\]
where ${\rm{tri}}(\cdot)$ denotes a triangular waveform and $\xi[n]$ is a uniform noise in the range $[-1, 1]$. Each signal is composed of $L = 30,000$ samples. The mixing matrix is a $5 \times 5$ Hilbert matrix, which is extremely ill-conditioned. All simulations have been performed by MATLAB 2015a, on an Intel Core i7 3.10 GHz processor at 64 bit with 8 GB of RAM. Parameters of the algorithms have been found heuristically.

\begin{figure}
\centering
\includegraphics[width=0.6\columnwidth]{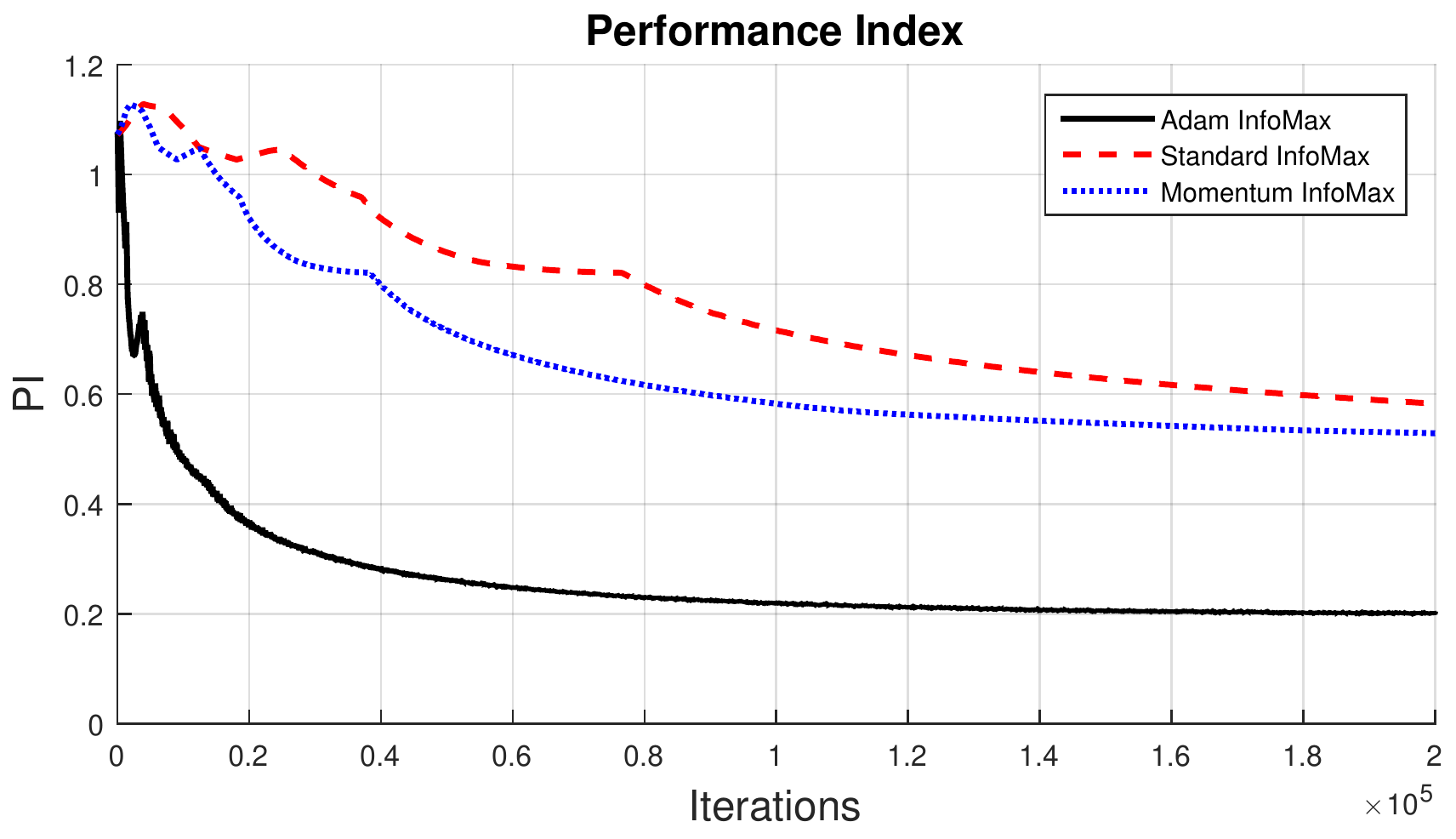}
\caption{Performance Index (PI) of the first proposed experiment.}
\label{fig:ex1}
\end{figure}

We perform separation by the Adam modification of the standard InfoMax algorithm, with gradient in \eqref{eq:entropy_gradient}. We use a block length of $B = 30$ samples (hence $N_B = \left\lfloor L/B \right\rfloor = 1,000$), while the other parameters are set as: $N_{ep} = 200$, $\beta_1 = 0.5$, $\beta_2 = 0.75$, $\varepsilon = 10^{-8}$, $\eta = 0.001$ and the learning rate of the standard InfoMax and the Momentum InfoMax to $\mu = 5 \cdot 10^{-5}$. Performance in terms of the PI in \eqref{eq:PI} is reported in Figure \ref{fig:ex1}, that clearly shows the effectiveness of the proposed idea.

\noindent A second and third experiments are performed on speech audio signals sampled at 8 kHz. Each signal is composed of $L = 30,000$ samples. In the second experiment a male and a female speech are mixed with a $2 \times 2$ random matrix with entries uniformly distributed in the interval $[-1, 1]$ while in the third one, two male and two female speeches are mixed with a $4 \times 4$ ill-conditioned Hilbert matrix. In addition, an additive white noise with 30 dB of SNR is added to the mixtures in both cases.

\begin{figure}
\centering
\subfloat[][]{\includegraphics[width=0.5\columnwidth]{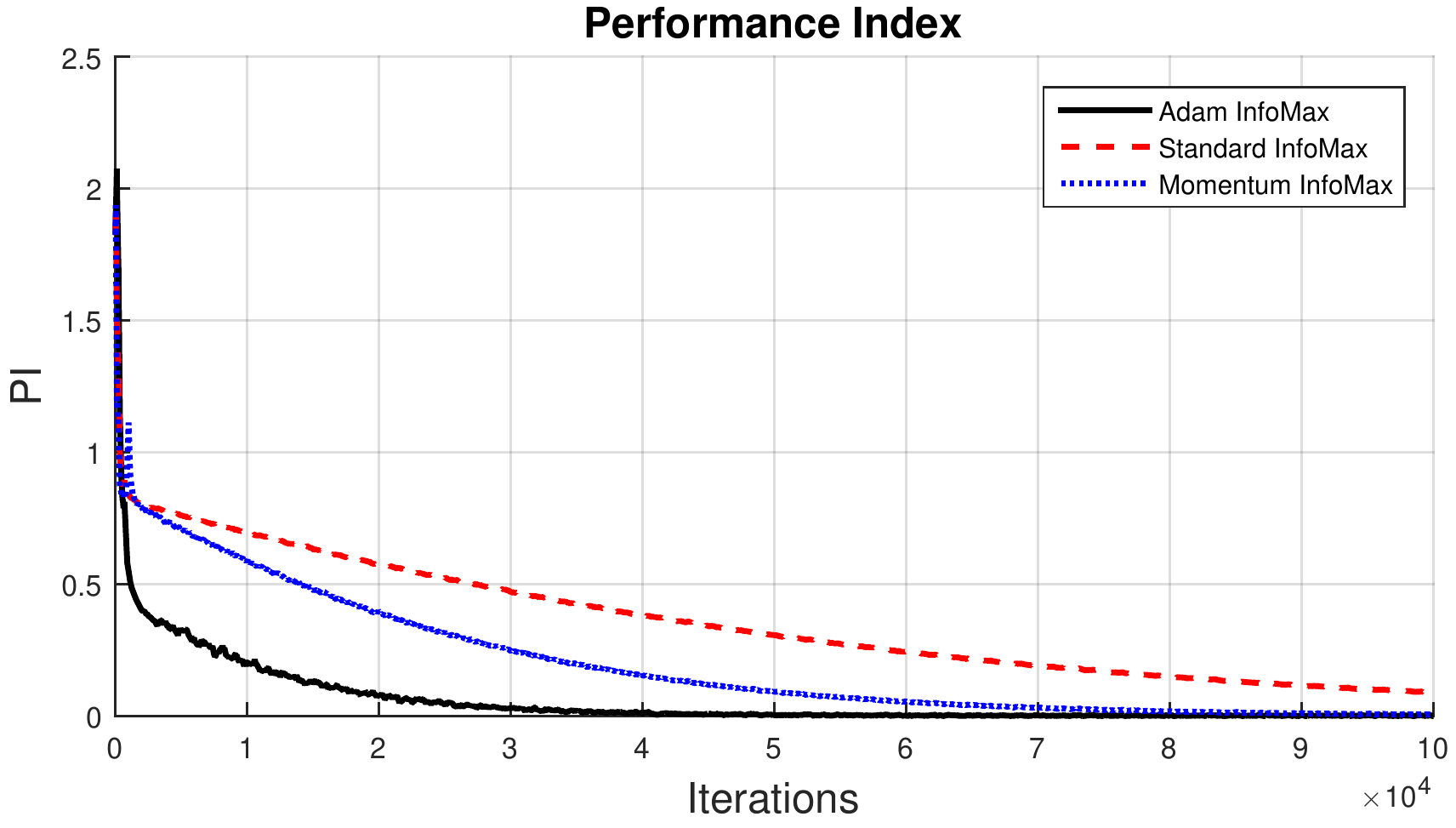}\label{fig:PI1}}%
\subfloat[][]{\includegraphics[width=0.5\columnwidth]{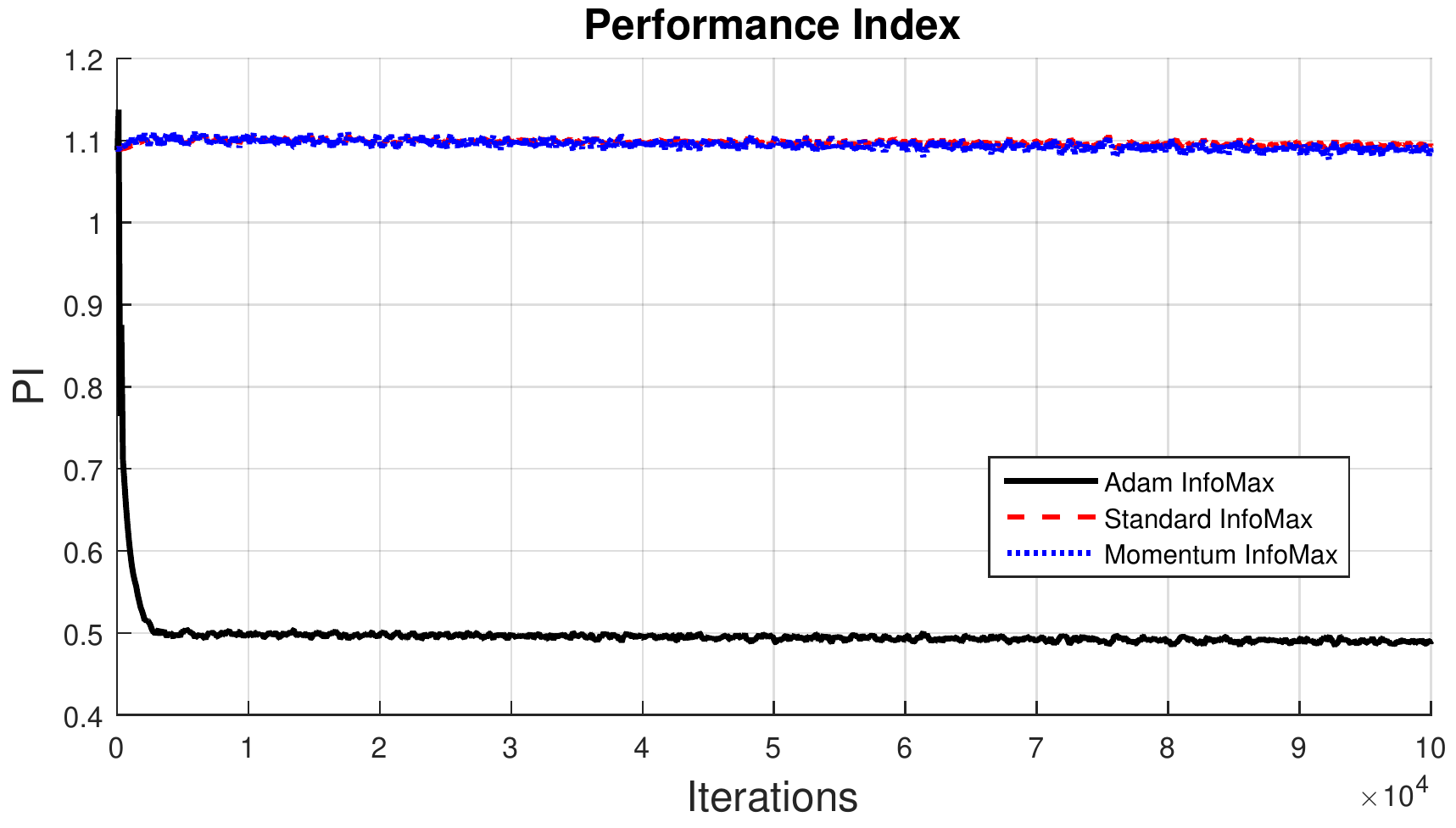}\label{fig:PI3}}
\caption{Performance Index (PI) of the: second (a) and third (b) proposed experiment.}%
\label{fig:ex23}
\end{figure}

We perform separation by the Adam modification of the NG InfoMax algorithm, with gradient in \eqref{eq:entropy_natural_gradient}. We use a block length of $B = 30$ samples, while the other parameters are set as: $N_{ep} = 100$, $\beta_1 = 0.9$, $\beta_2 = 0.999$, $\varepsilon = 10^{-8}$, $\eta = 0.001$ and the learning rate of the standard InfoMax and the Momentum InfoMax to $\mu = 0.001$. Performances in terms of the PI, for the second and third experiments, are reported in Figures \ref{fig:PI1} and \ref{fig:PI3}, respectively, that clearly show also in these cases the effectiveness of the proposed idea. In particular, Figure \ref{fig:PI3} confirms that the separation obtained by using the Adam InfoMax algorithm in the third experiment is quite satisfactory, while the standard and the Momentum InfoMax give worse solutions.

\begin{figure}
\centering
\includegraphics[width=0.6\columnwidth]{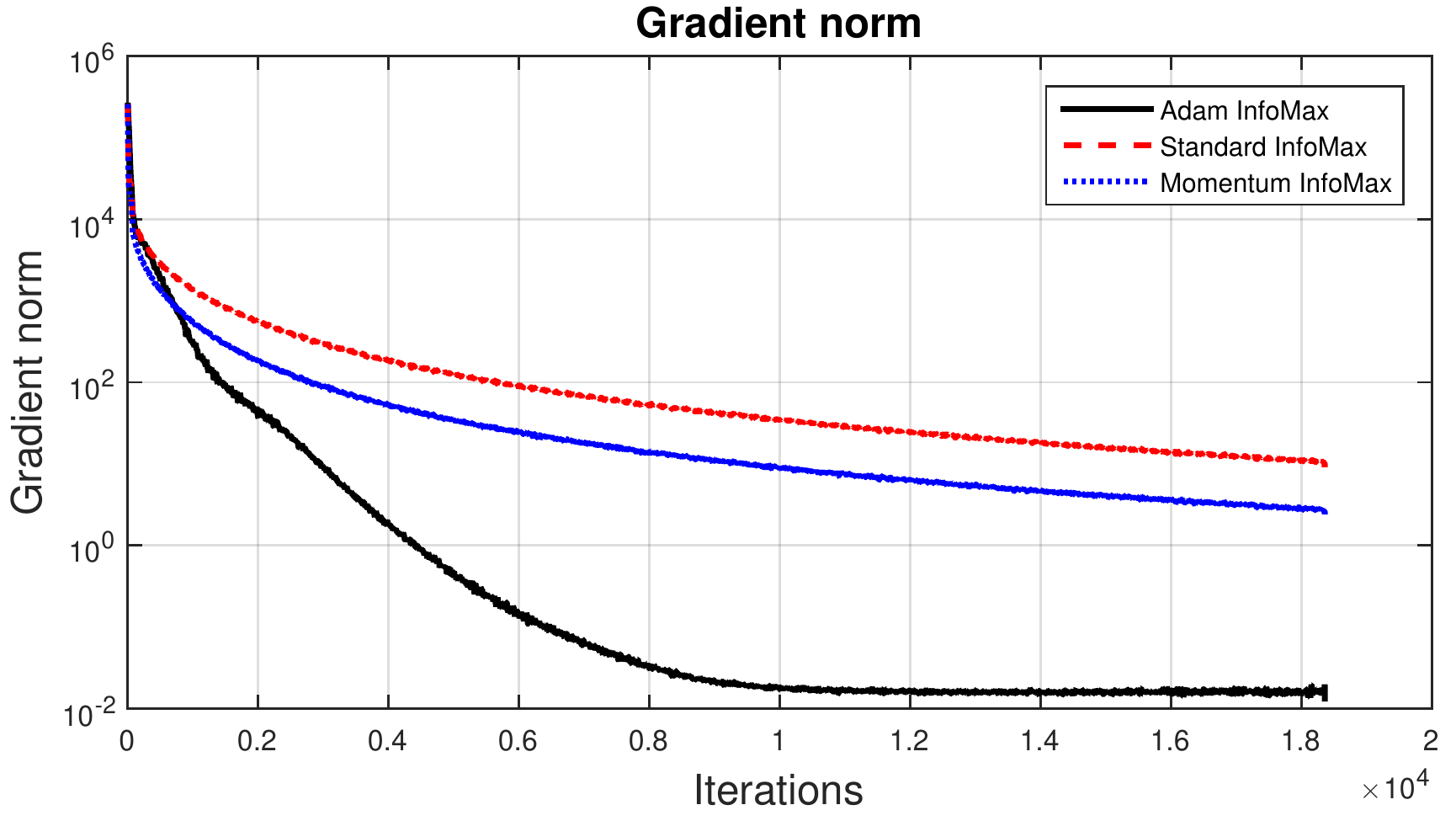}
\caption{Norm of the gradient of the cost function in the fourth proposed experiment.}
\label{fig:ex4}
\end{figure}

Finally, a last experiment is performed on real data. We used an EEG signal recorded according the 10-20 system, consisting of 19 signals with artifacts. ICA is a common approach to deal with the problem of artifact removal from EEG \cite{Inuso2007}. We use a block length of $B = 30$ samples, while the other parameters are set as: $N_{ep} = 270$, $\beta_1 = 0.9$, $\beta_2 = 0.999$, $\varepsilon = 10^{-8}$, $\eta = 0.01$ and the learning rate of the standard InfoMax and the Momentum InfoMax to $\mu = 10^{-6}$. Since we used real data and the mixing matrix $\mathbf{A}$ is not available, the PI cannot be evaluated. Hence, we decided to evaluate the performance by the norm of the gradient of the cost function. As it can be seen from Figure \ref{fig:ex4}, also in this case the Adam InfoMax algorithm achieves better results in a smaller number of iterations with respect to the compared algorithms.

\section{Conclusions}
\label{sec:conclusions}

In this paper a modified InfoMax algorithm for the blind separation of independent sources, in a linear and instantaneous environment, has been introduced. The proposed approach is based on a novel and advanced stochastic optimization method known as Adam and it can benefit from the excellent properties of the Adam approach. In particular, it is easy to implement, computationally efficient, and it is well suited when the number of sources is high and bad-scaled, the mixing matrix is close to be ill-conditioned and some additive noise is considered. Some experimental results, evaluated in terms of the Amari Performance Index and compared with other state-of-the-art approaches, have shown the effectiveness of the proposed approach.


\end{document}